\documentclass[11pt]{article}
\usepackage{acl2016}
\usepackage{times}
\usepackage{latexsym}

\usepackage{url}

\usepackage{graphicx}
\usepackage{amssymb,amsmath,bm}
\usepackage{textcomp}
\usepackage{multirow}
\usepackage{dsfont}
\usepackage{color}
\usepackage[normalem]{ulem}
\usepackage{rotating}
\usepackage{booktabs}
\usepackage{makecell}
\usepackage{comment}
\usepackage{amsmath,amsfonts,amssymb}
\usepackage{tikz}
\usepackage{verbatim}
\usepackage{caption}
\usepackage{booktabs}
\usepackage{ctable}
\usepackage{makecell}
\usepackage{array} % for defining a new column type
\usepackage{varwidth} %for the varwidth minipage environment
\usepackage{stmaryrd}
\usepackage{hyperref}

\aclfinalcopy % Uncomment this line for the final submission
\def\aclpaperid{142} %  Enter the acl Paper ID here

\title{On-line Active Reward Learning for Policy Optimisation\\ in Spoken Dialogue Systems} 

\author{
Pei-Hao Su, Milica Ga{\v s}i{\' c}, Nikola Mrk{\v s}i{\' c}, Lina Rojas-Barahona, \\ {\bf Stefan Ultes, David Vandyke, Tsung-Hsien Wen and Steve Young} \\
Department of Engineering,
University of Cambridge,
Cambridge, UK\\
\texttt{\{phs26, mg436, nm480, lmr46, su259, djv27, thw28, sjy\}@cam.ac.uk} \\
}
\date{}

\begin{document}

\maketitle

\begin{abstract}

The ability to compute an accurate reward function is essential for optimising a dialogue policy via reinforcement learning. In real-world applications,
using explicit user feedback as the reward signal is often unreliable and costly to collect.
This problem can be mitigated if the user's intent is known in advance or data is available to pre-train a task success predictor off-line.
In practice neither of these apply for most real world applications. 
Here we propose an on-line learning framework whereby the dialogue policy is jointly trained  alongside the reward model via active learning with a Gaussian process model. This Gaussian process operates on a continuous space dialogue representation generated in an unsupervised fashion using a recurrent neural network encoder-decoder.
The experimental results demonstrate that the proposed framework is able to significantly reduce data annotation costs and mitigate noisy user feedback in dialogue policy learning.

\end{abstract}

\section{Introduction} \label{sec:intro}

Spoken Dialogue Systems (SDS) allow human-computer interaction using natural speech. They can be broadly divided into two categories: chat-oriented systems which aim to converse with users and provide reasonable contextually relevant responses \cite{vinyals2015neural,serban2015hierarchical},  and task-oriented systems designed to assist users to achieve specific goals (e.g. find hotels, movies or bus schedules) \cite{KTD,POMDP-review}. The latter are typically designed according to a structured {\it ontology} (or a database {\it schema}), which defines the domain that the system can talk about. Teaching a system how to respond appropriately in a task-oriented SDS is non-trivial. This {\it dialogue management} task is often formulated as a manually defined  dialogue flow that directly determines the quality of interaction. More recently, dialogue management has been formulated as a reinforcement learning (RL) problem which can be automatically optimised \cite{levin1997stochastic,roy2000spoken,POMDP_williams,POMDP-review}. In this framework, the system learns by a {\it trial and error} process governed by a potentially delayed learning objective defined by a {\it reward function}.

\begin{figure}[t]
\centerline{\includegraphics[scale=0.23]{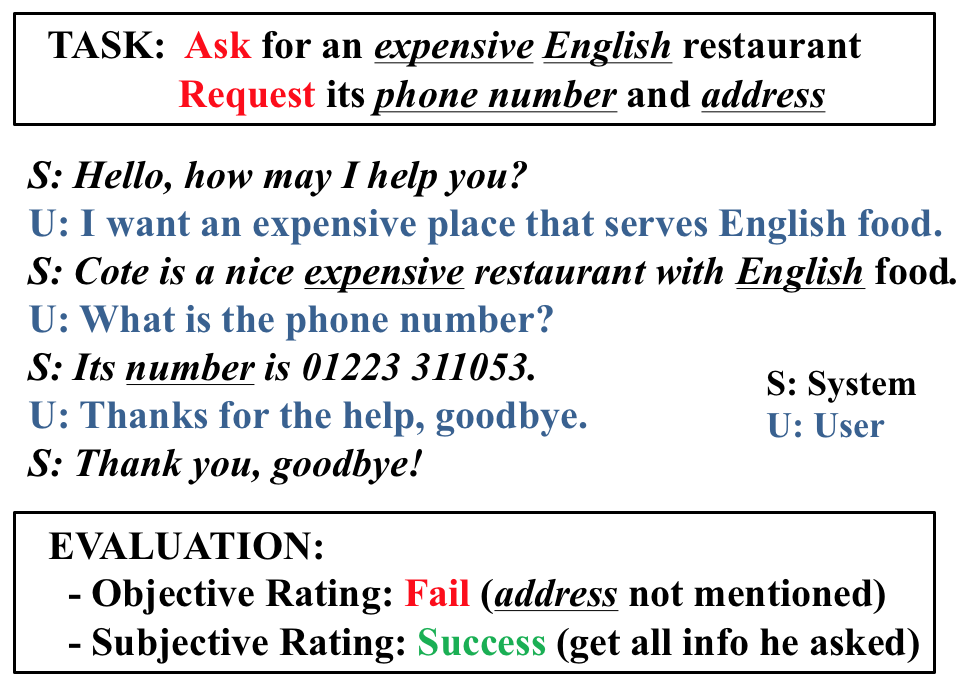}}
\caption{{An example of a task-oriented dialogue with a pre-defined task and the evaluation results.}} 
\label{fig:dialogueExample}
\end{figure}

A typical approach to defining the reward function in a task-oriented dialogue system is to apply a small per-turn penalty to encourage short dialogues 
and to give a large positive reward at the end of each successful interaction. 
Figure \ref{fig:dialogueExample} is an example of a dialogue task which is typically set for users who are being paid to converse with the system.
When users are primed with a specific task to complete, dialogue success can be determined from subjective user ratings (\textit{Subj}), or an objective measure (\textit{Obj}) based on whether or not the pre-specified task was completed
\cite{PARADISE,milica_icassp13}. However,  prior knowledge of the user's goal is not normally available in real situations, making the objective reward estimation approach impractical. 

Furthermore, objective ratings are inflexible and often fail as can be seen from Figure \ref{fig:dialogueExample}, if the user does not strictly follow the task. This results in a mismatch between the \textit{Obj} and \textit{Subj} ratings. However, relying on subjective ratings alone is also problematic since crowd-sourced subjects frequently give inaccurate responses and real users are often unwilling to extend the interaction in order to give feedback,  resulting in unstable learning \cite{zhao2011incremental,milica_real_users}. 
In order to filter out incorrect user feedback, \newcite{milica_icassp13} used only dialogues for which $\textit{Obj} = \textit{Subj}$. Nonetheless, this is inefficient and not feasible anyway in most real-world tasks 
where the user's goal is generally unknown and difficult to infer.

In light of the above, \newcite{Su_2015} proposed learning a neural network-based \textit{Obj} estimator from off-line simulated dialogue data. This removes the need for the \textit{Obj} check during on-line policy learning and the resulting policy is as effective as  one trained with dialogues using the $\textit{Obj} = \textit{Subj}$ check. However, a user simulator will only provide a rough approximation of real user statistics and developing a user simulator is a costly process \cite{schatzmann2006survey}. 

To deal with the above issues, this paper describes an on-line active learning method in which users are asked to provide feedback on whether the dialogue was successful or not.  However, active learning is used to limit requests for feedback to only those cases where the feedback would be useful, and also a noise model is introduced to compensate for cases where the user feedback is inaccurate.
A Gaussian process classification (GPC) model is utilised to robustly model the uncertainty presented by the noisy user feedback.
Since GPC operates on a fixed-length observation space and dialogues are of variable-length, a recurrent neural network (RNN)-based embedding function is used to provide fixed-length dialogue representations. In essence, the proposed method learns a dialogue policy and a reward estimator on-line from scratch, and is directly applicable to real-world applications. 

The rest of the paper is organised as follows. The next section gives an overview of  related work. The proposed framework is then
described in \S \ref{sec:models}. This consists of the policy learning algorithm, the creation of the dialogue embedding function and the active reward model trained from real user ratings. 
In \S \ref{sec:exp}, the proposed approach is evaluated in the context of an application providing restaurant information in Cambridge, UK. We first give an in-depth analysis of the dialogue embedding space. The results of the active reward model when it is trained together with a dialogue policy on-line with real users are then presented.
Finally, our conclusions are presented in \S \ref{sec:conclude}.

\section{Related Work}

Dialogue evaluation has been an active research area since late 90s. \newcite{PARADISE} proposed the PARADISE framework, where a linear function of task completion and various dialogue features such as dialogue duration were used to infer user satisfaction. This measure was later used as a reward function for learning a dialogue policy \cite{rieser2011learning}. However, as noted, task completion is rarely available when the system is interacting with real users and also concerns have been raised regarding the theoretical validity of the model \cite{larsen_2003}. 

Several approaches have been adopted for learning a dialogue reward model given a corpus of annotated dialogues. \newcite{yang2012predicting} used collaborative filtering to infer user preferences. The use of reward shaping has also been investigated in \cite{el2014task,svgm15} to enrich the reward function in order to speed up dialogue policy learning. Also, \newcite{2015su03} demonstrated that there is a strong correlation between expert's user satisfaction ratings and dialogue success.
However, all these methods assume the availability of reliable dialogue annotations such as expert ratings, which in practice are hard to obtain.

One effective way to mitigate the effects of annotator error is to obtain multiple ratings for the same data and several methods have been developed to guide the annotation process with uncertainty models \cite{dai2013pomdp,lin2014re}. Active learning is particularly useful for determining when an annotation is needed \cite{settles2010active,ZhangC15c}. It is often utilised using Bayesian optimisation approaches \cite{brochu2010tutorial}.
Based on this, \newcite{Daniel_RSS_2014} exploited a pool-based active learning method for a robotics application. They queried the user for feedback on the most informative sample collected so far and showed the effectiveness of this method.  

Rather than explicitly defining a reward function, inverse RL (IRL) aims to recover the underlying reward from demonstrations of good behaviour and then learn a policy which maximises the recovered reward \cite{russell1998learning}. 
IRL was first introduced to SDS in \cite{paek2008automating}, where the reward was inferred from human-human dialogues to mimic the behaviour observed in a corpus. IRL has also been studied in a Wizard-of-Oz (WoZ) setting \cite{boularias2010learning,rojasbarahona:hal-01002361}, where typically a human expert served as the dialogue manager to select each system reply based on the speech understanding output at different noise levels. 
However, this approach is costly and there is no reason to suppose that a human wizard is acting optimally, especially at high noise levels.

Since humans are better at giving relative judgements than absolute scores, another related line of research has focused on preference-based approaches to RL \cite{cheng2011preference}. In \cite{sugiyama2012preference},  users were asked to provide rankings between pairs of dialogues. However, this is also costly and does not scale well in real applications.

\section{Proposed Framework} \label{sec:models}

\begin{figure*}[t]
\centerline{\includegraphics[scale=0.5]{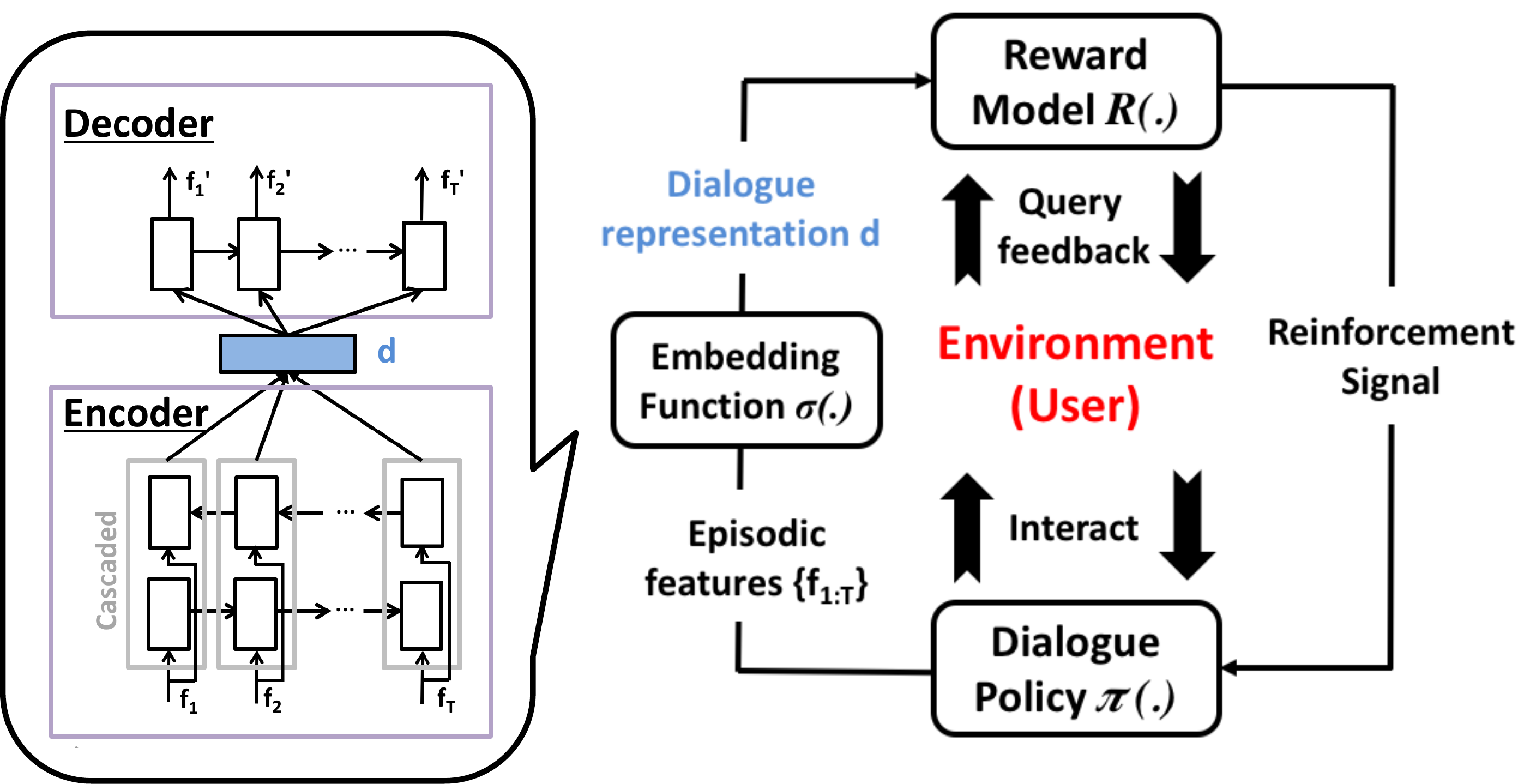}}
\caption{Schematic of the system framework. The three main system components dialogue policy, dialogue embedding creation, and reward modelling based on user feedback, are described in \S \ref{sec:models}.} 
\label{fig:framework}
\end{figure*}

The proposed system framework is depicted in Figure \ref{fig:framework}.  It is divided into three main parts:
a dialogue policy, a dialogue embedding function, and an active reward model of user feedback. 
When each dialogue ends, a set of turn-level features $\mathbf{f_t}$ is extracted and fed into an embedding function $\sigma$ to obtain a fixed-dimension dialogue representation $\mathbf{d}$ that serves as the input space of the reward model $R$. This reward is modelled as a Gaussian process which for every input point provides an estimate of task success along with a measure of the estimate uncertainty. Based on this uncertainty, $R$ decides whether to query the user for feedback or not.  It then returns a reinforcement signal to update the dialogue policy $\pi$, which is trained using the GP-SARSA algorithm \cite{GPRL}. GP-SARSA also deploys Gaussian process estimation to provide an on-line sample-efficient reinforcement learning algorithm capable of bootstrapping estimates of sparse value functions from minimal numbers of samples (dialogues). The quality of each dialogue is defined by its cumulative reward, where each dialogue turn incurs a small negative reward (-1) and the final reward of either 0 or 20 
depending on the estimate of task success are provided by the reward model.

Note that the key contribution here is to learn the noise robust reward model and the dialogue policy simultaneously on-line, using the user as a `supervisor'.  Active learning is not an essential component of the framework but highly desirable in practice to minimise the impact of the supervision burden on users. 
The use of a pre-trained embedding function is a sub-component of the proposed approach and is trained off-line on corpus data rather than manually designed here.

\subsection{Unsupervised Dialogue Embeddings} \label{sec:de}

In order to model user feedback over dialogues of varying length, an embedding function is used to map each dialogue into a fixed-dimensional continuous-space.
The use of embedding functions has recently gained attention especially for word representations, and has boosted  performance on several natural language processing tasks \cite{mikolov2013distributed,turian2010word,levy2014neural}. 
Embedding has also been successfully applied to machine translation (MT) where it enables 
varying-length phrases to be mapped to fixed-length vectors using an RNN Encoder-Decoder \cite{cho2014learning}.
Similar to MT, dialogue embedding enables variable length sequences of utterances to be mapped into an appropriate fixed-length vector.    Although embedding is used here to create a fixed-dimension input space for the GPC-based task success classifier, it should be noted that it potentially facilitates a variety of other downstream tasks which depend on classification or clustering.

The model structure of the embedding function is described on the left of Figure \ref{fig:framework}, where the episodic turn-level features $\mathbf{f_t}$ are extracted from a dialogue 
and serve as the input features to the encoder. 
In our proposed model, the encoder is a Bi-directional Long Short-Term Memory network (BLSTM) \cite{hochreiter1997long,graves2013hybrid}. The LSTM is a Recurrent Neural Network (RNN) with gated recurrent units introduced to alleviate the vanishing gradient problem. The BLSTM encoder takes into account the sequential information from both directions of the input data, computing the {\it forward} hidden sequences $ \overrightarrow{\mathbf{h}}_{1:T}$ and the {\it backward} hidden sequences $\overleftarrow{\mathbf{h}}_{T:1}$ while iterating over all input features $\mathbf{f_{t}}$, $t=1,...,T$:
\begin{align*}
\overrightarrow{\mathbf{h_t}} = {LSTM}(\mathbf{f_{t}},\overrightarrow{\mathbf{h}}_{t-1}) \\
\overleftarrow{\mathbf{h_t}}  = {LSTM}(\mathbf{f_{t}},\overleftarrow{\mathbf{h}}_{t+1})
\end{align*}
where ${LSTM}$ denotes the activation function.
The dialogue representation $\mathbf{d}$ is then calculated as the average over all hidden sequences:

\begin{equation}
\mathbf{d} = \frac{1}{T}{\sum_{t=1}^{T} \mathbf{h_t}}
\end{equation}
where $\mathbf{h_t} = [\overrightarrow{\mathbf{h_t}};\overleftarrow{\mathbf{h_t}}]$ is the concatenation of the two directional hidden sequences.

Given the dialogue representation $\mathbf{d}$ output by the encoder, the decoder is a forward LSTM that takes $\mathbf{d}$ as its input for each turn $t$ to produce the sequence of features $\mathbf{f'}_{1:T}$.

The training objective of the encoder-decoder minimises the mean-square-error (MSE) between the prediction $\mathbf{f'}_{1:T}$ and the output $\mathbf{f}_{1:T}$ (which is also the input):

\begin{equation}
MSE = \frac{1}{N}{\sum_{i=1}^{N} \sum_{t=1}^{T} ||\mathbf{f}_{t} - \mathbf{f}_{t}'||^2}
\end{equation}
where $N$ is the number of training dialogues and $||\cdot||^2$ denotes the $l^2$-norm. Since all the functions used in the encoder and decoder are differentiable, stochastic gradient decent (SGD) can be used to train the model.

The dialogue representations generated by this LSTM-based unsupervised embedding function are then used as the observations for the reward model described in the next section \ref{sec:arl}. 

\subsection{Active Reward Learning} \label{sec:arl}

A Gaussian process is a Bayesian non-parametric model that can be used for regression or classification~\cite{rasmussen2006gaussian}. It is particularly appealing since it can learn from a small number of observations by exploiting the correlations defined by a \emph{kernel function} and it provides a measure of uncertainty of its estimates. In the context of spoken dialogue systems it has been successfully used for RL policy optimisation~\cite{GPRL,casanueva2015} and  IRL reward function regression~\cite{kim2014inverse}.

Here we propose modelling dialogue success as a Gaussian process (GP). This involves estimating the probability $p(y| \mathbf{d}, \mathcal{D})$ that the task was successful given the current dialogue representation $\mathbf{d}$ and the pool $\mathcal{D}$ containing previously classified dialogues. We pose this as a classification problem where the rating is a binary observation $y\in\{-1, 1\}$ that defines failure or success.
The observations $y$ are considered to be drawn from a Bernoulli distribution with a success probability $p(y=1| \mathbf{d}, \mathcal{D})$. The probability is related to a latent function $f(\mathbf{d}|\mathcal{D}): \mathcal{R}^{dim(\mathbf{d})} \to \mathcal{R}$ that is mapped to a unit interval by a {\it probit} function $p(y=1| \mathbf{d}, \mathcal{D}) = \phi(f(\mathbf{d}|\mathcal{D}))$, where $\phi$ denotes the
cumulative density function of the standard Gaussian distribution.

The latent function is given a GP prior: $f(\mathbf{d}) \sim \mathcal{GP}(m(\mathbf{d}), k(\mathbf{d}, \mathbf{d'}))$, where ${m(\cdot)}$ is the mean function and ${k(\cdot,\cdot)}$ the covariance function (kernel). Here the stationary squared exponential kernel ${k_{SE}}$ is used. It is also combined with a white noise kernel ${k_{WN}}$ in order to account for the ``noise'' in users' ratings:

\vspace{-5mm}
\begin{equation}
{k(\mathbf{d}, \mathbf{d'})} = p^2 \exp(-\frac{||\mathbf{d} - \mathbf{d'}||^2}{2l^2}) + \sigma^{2}_n
\label{eq:kernel}
\end{equation}
where the first term denotes ${k_{SE}}$ and the second term ${k_{WN}}$. 

The {\it hyper-parameters} $p, l, \sigma_n$ can be adequately optimised by maximising the marginal likelihood using a gradient-based method \cite{chen2015hyper}.   Since $\phi(\cdot)$ is not Gaussian, the resulting posterior probability $p(y=1| \mathbf{d}, \mathcal{D})$ is analytically intractable. So instead an approximation method, expectation propagation (EP), was used \cite{nickisch2008approximations}.

Querying the user for feedback is costly and may impact negatively on the user experience.   This impact can be reduced by using active learning informed by the uncertainty estimate of the GP model \cite{kapoor2007active}. This ensures that user feedback is only sought when the model is uncertain about its current prediction. 
For the current application, an on-line (stream-based) version of 
active learning is required.

\begin{figure}[t]
\centerline{\includegraphics[scale=0.24]{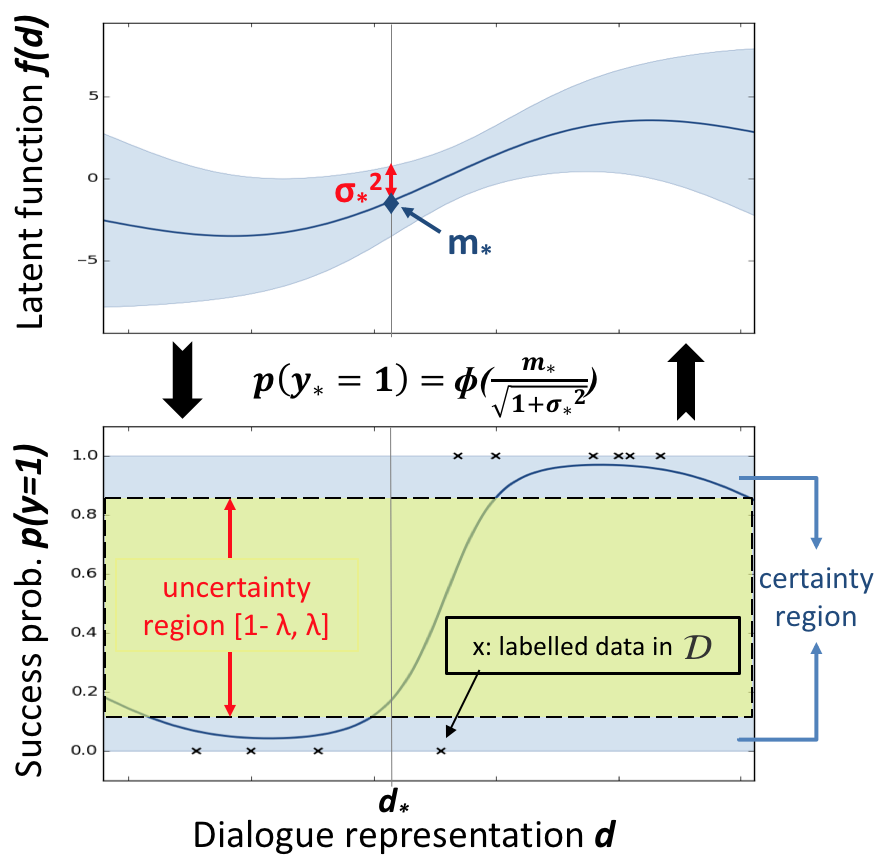}}
\caption{{1-dimensional example of the proposed GP active reward learning model.}} 
\label{fig:gp}
\end{figure}

An illustration of a 1-dimensional example is shown in Figure \ref{fig:gp}. Given the labelled data $\mathcal{D}$,
the predictive posterior mean $\mu_{\ast}$ and posterior variance ${\sigma}^2_{\ast}$ of the latent value $f(\mathbf{d_{\ast}})$ for the current dialogue representation $\mathbf{d_{\ast}}$ can be calculated. Then a threshold interval $[ 1-\lambda, \lambda ]$ is set on the predictive success probability $p(y_{\ast}=1| \mathbf{d_{\ast}}, \mathcal{D}) = \phi(\mu_{\ast}/\sqrt{1+{\sigma}^2_{\ast}} )$ to decide whether this dialogue should be labelled or not. The decision boundary implicitly considers both the posterior mean as well as the variance.

When deploying this reward model in the proposed framework, a GP with a zero-mean prior for $f$ is initialised and $\mathcal{D} = \{\}$. After the dialogue policy $\pi$ completes each episode with the user, the generated dialogue turns are transformed into the dialogue representation $\mathbf{d} = \sigma( \mathbf{f}_{1:T})$ using the dialogue embedding function $\sigma$.
Given $\mathbf{d}$, the predictive mean and variance of $f(\mathbf{d}|\mathcal{D})$ are determined, and the reward model decides whether or not it should seek user feedback based on the threshold $\lambda$ on $\phi(f(\mathbf{d}|\mathcal{D}))$. If the model is uncertain, the user's feedback on the current episode $\mathbf{d}$ is used to update the GP model and to generate the reinforcement signal for training the policy $\pi$; otherwise the predictive success rating from the reward model is used directly to update the policy. This process takes place after each dialogue.

\section{Experimental results} \label{sec:exp}
The target application is a live telephone-based spoken dialogue system providing restaurant information for the Cambridge (UK) area. The domain consists of approximately 150 venues each having 6 slots (attributes) of which 3 can be used by the system to constrain the search (food-type, area and price-range) and the remaining 3 are informable properties (phone-number, address and postcode) available once a required database entity has been found. 

The shared core components of the SDS common to all experiments comprise a HMM-based recogniser, a confusion network (CNet) semantic input decoder \cite{CNET}, the BUDS belief state tracker \cite{BUDS} that factorises the dialogue state using a dynamic Bayesian network, and a template based natural language generator to map system semantic actions into natural language responses to the user. 
All policies were trained using the GP-SARSA algorithm and the summary action space of the RL policy contains 20 actions. 

The reward given to each dialogue was set to $20\times\mathds{1}_{success}- N$, where $N$ is the dialogue turn number and $\mathds{1}$ is the indicator function for dialogue success, which is determined by different methods as described in the following section. These rewards constitute the reinforcement signal used for policy learning.

\subsection{Dialogue representations} \label{sec:dr}

The LSTM Encoder-Decoder model described in \S \ref{sec:de} was used
to generate an embedding $\mathbf{d}$ for each dialogue.
For each dialogue turn that contains a user's utterance and a system's response, a feature vector $\mathbf{f}$ of size $74$ was extracted \cite{Vandyke_2015}.
This vector consists of the concatenation of the most likely user intention determined by the semantic decoder, the distribution over each concept of interest defined in the ontology, a one-hot encoding of the system's reply action, and the turn number normalised by the maximum number of turns (here 30).
This feature vector was used as the input and the target for the LSTM Encoder-Decoder model, where the training objective was to minimise the MSE of the reconstruction loss. 

The model was implemented using the Theano library \cite{bergstra+al:2010-scipy,Bastien-Theano-2012}. A corpus consisting of 8565, 1199 and 650 real user dialogues in the Cambridge restaurant domain was used for training, validation and testing respectively. This corpus was collected via the Amazon Mechanical Turk (AMT) service, where paid subjects interacted with the dialogue system. The sizes of  $\overrightarrow{\mathbf{h_t}}$ and $\overleftarrow{\mathbf{h_t}}$ in the encoder and the hidden layer in the decoder were all 32, resulting in $dim(\mathbf{h_t})$ = $dim(\mathbf{d})$ = 64. SGD per dialogue was used during backpropagation to train each model. In order to prevent over-fitting, {\it early stopping} was applied based on the held-out validation set.

\begin{figure}[t]
\centerline{\includegraphics[scale=0.43]{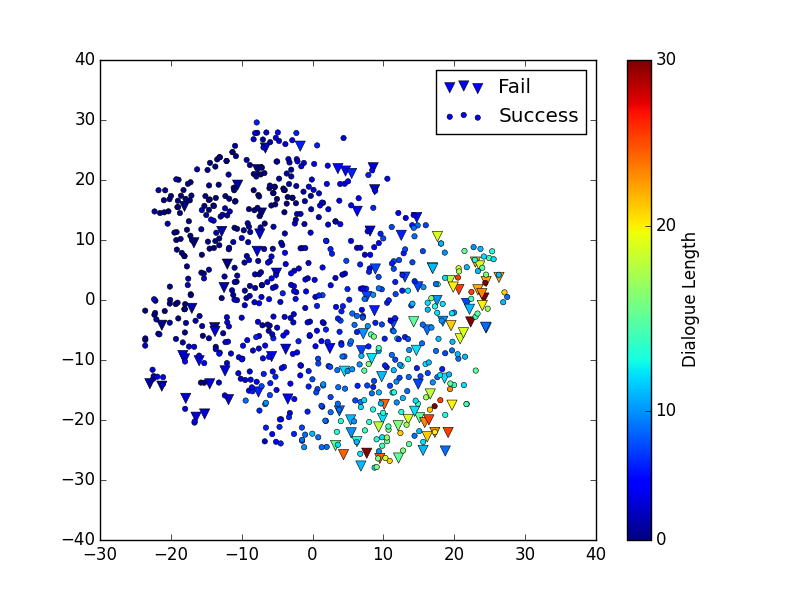}}
\caption{{t-SNE visualisation on the unsupervised dialogue representation of the real user data in the Cambridge restaurant domain. Labels are the subjective ratings from the users.}} 
\label{fig:tsne}
\end{figure}

In order to visualise the impact of the embeddings, the dialogue representations of all the 650 test dialogues were transformed by the embedding function in Figure \ref{fig:tsne} and reduced to two dimensions using t-SNE \cite{van2008visualizing}. For each dialogue sample, the shape indicates whether or not the dialogue was successful, and the colour indicates the
length of the dialogue (maximum 30 turns). 

From the figure we can clearly see the colour gradient from the top left (shorter dialogues) to the bottom right (longer dialogues) for the positive \textit{Subj} labels. This shows that dialogue length was one of the prominent features in the dialogue representation $\mathbf{d}$. 
It can also be seen that the longer failed dialogues (more than 15 turns) are located close to each other, mostly at the bottom right. On the other hand, there are other failed dialogues which are spread throughout the cluster. 
We can also see that the successful dialogues were on average shorter than 10 turns, which is consistent with the claim that users do not engage in longer dialogues with well-trained task-oriented systems.

This visualisation shows the potential of the unsupervised dialogue embedding since the transformed dialogue representations appear to be correlated with dialogue success in the majority of cases. 
For the purpose of GP reward modelling, this LSTM Encoder-Decoder embedding function appears therefore to be suitable for extracting an adequate fixed-dimension dialogue representation.

\subsection{Dialogue Policy Learning} \label{sec:dpl}

Given the well-trained dialogue embedding function, the proposed GP reward model operates on this input space. The system was implemented using the GPy library \cite{gpy2014}. Given the predictive success probability of each newly seen dialogue, the threshold $\lambda$ for the uncertainty region was initially set to 1 to encourage label querying and annealed to 0.85 for the first 50 collected dialogues and then set to 0.85 thereafter.

Initially, as each new dialogue was added to the training set,
the hyper-parameters that defined the structure of the kernels mentioned in Eqn. \ref{eq:kernel} were optimised to minimise the negative log marginal likelihood using conjugate gradient ascent \cite{rasmussen2006gaussian}. To prevent {\it overfitting}, after the first 40 dialogues, these hyper-parameters were only re-optimised after every batch of 20  dialogues.

\begin{figure}[t]
\centerline{\includegraphics[scale=0.5]{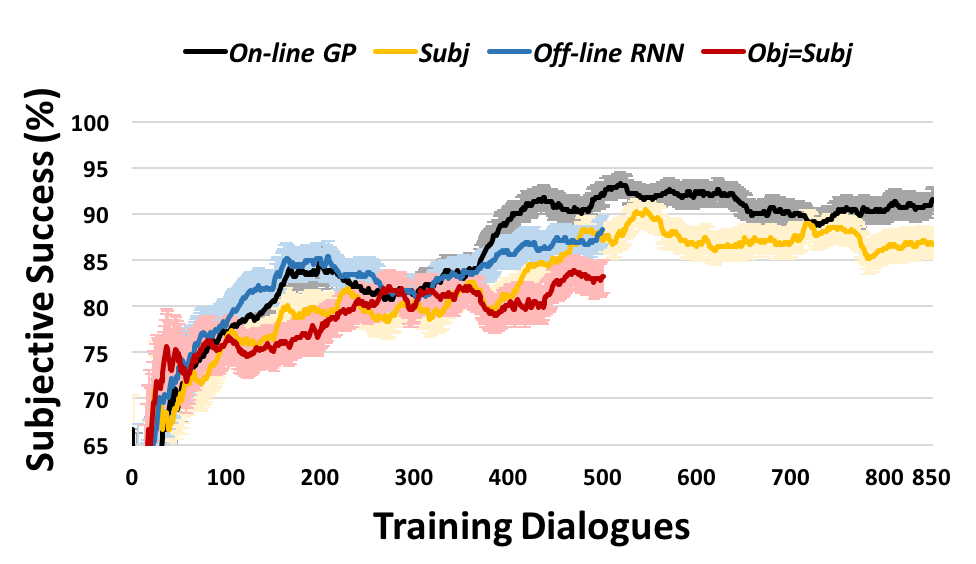}}
\caption{{Learning curves showing subjective success as a function of the number of training dialogues used during on-line policy optimisation. The {\it on-line GP}, \textit{Subj}, \textit{off-line RNN} and \textit{Obj}=\textit{Subj} systems are shown as black, yellow, blue, and red lines. 
The light-coloured areas are one standard error intervals.}} 
\label{fig:Subjsuc}
\end{figure}

To investigate the performance of the proposed {\it on-line GP} policy learning,
three other contrasting systems were also tested. 
Note that the hand-crafted system is not compared since it does not scale to larger domains and is sensitive to speech recognition errors. In each case, the only difference was the method used to compute the reward:

\begin{itemize}
\item the \textit{Obj}=\textit{Subj} system which uses prior knowledge of the task to only use training dialogues for which the user's subjective assessment of success is consistent with the objective assessment of success as in
\cite{milica_icassp13}.
\item the \textit{Subj} system which directly optimises the policy using only the user assessment of success whether accurate or not.
\item the \textit{off-line RNN} system that uses 1K simulated data and the corresponding \textit{Obj} labels to train an RNN success estimator as in \cite{Su_2015}. %Cross-entropy loss function and sigmoid activations were adopted.
\end{itemize}

For the \textit{Subj} system rating, in order to focus solely on the performance of the policy rather than 
other aspects of the system such as the fluency of the reply sentence, users were asked to rate dialogue success by answering the following question:
{\it Did you find all the information you were looking for? }

All four of the above systems were trained with a total of 500 dialogues on-line by users recruited via the AMT service. 
Figure \ref{fig:Subjsuc} shows the on-line learning curve of the subjective success rating when during training. For each system, the moving average was calculated using a window of 150 dialogues. In each case, three distinct policies were trained and the results were averaged to reduce noise. 

\begin{figure}[t]
\centerline{\includegraphics[scale=0.5]{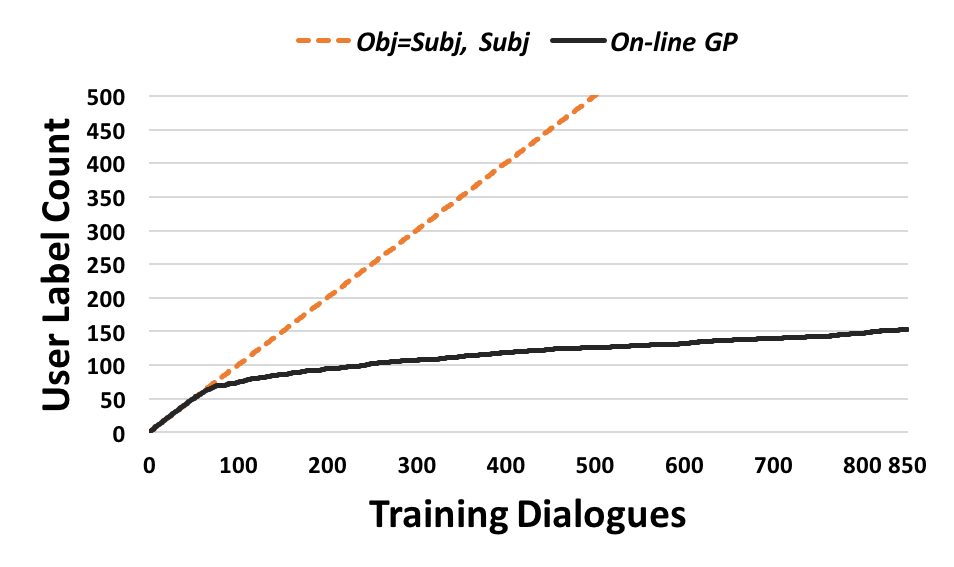}}
\caption{{The number of times each system queries the user for feedback during on-line policy optimisation as a function of the number of training dialogues. The orange line represents both the \textit{Obj}=\textit{Subj} and \textit{Subj} systems, and the black line represents the {\it on-line GP} system.}} 
\label{fig:acl}
\end{figure}

As can be seen, all four systems perform better than 80 \% subjective success rate after approximately 500 training dialogues. 
The \textit{Obj}=\textit{Subj} system is relatively poor compared to the others. This might be because users often report success even though the objective evaluation indicates failure. In such cases, the dialogue is discarded and not used for training.  As a consequence, the \textit{Obj}=\textit{Subj} system required approximately $700$ dialogues in order to obtain 500 which were useful, whereas all other systems made use of every dialogue. 

To investigate learning behaviour over longer spans, training for the {\it on-line GP} and the \textit{Subj} systems was extended to 850 dialogues.  As can be seen, performance in both cases is broadly flat.

Similar to the conclusions drawn in \cite{milica_real_users}, the \textit{Subj} system suffers from unreliable user feedback.  Firstly, as
in the \textit{Obj}=\textit{Subj} system, users forget the full requirements of the task and in particular, forget to ask for all required information.  Secondly, users give inconsistent feedback due to a lack of proper care and attention.  
From Figure \ref{fig:Subjsuc} it can be clearly seen that the {\it on-line GP} system consistently performed better than \textit{Subj} system, presumably, because its noise model mitigates the effect of inconsistency in user feedback.  Of course, unlike crowd-sourced subjects, real users might provide more consistent feedback, but nevertheless, some inconsistency is inevitable and the noise model offers the needed robustness.

The advantage of the {\it on-line GP} system in reducing the number of times that the system requests user feedback (i.e. the label cost) can be seen in Figure \ref{fig:acl}. The black curve shows the number of active learning queries triggered in the {\it on-line GP} system averaged across the three policies.  This system required only 150 user feedback requests to train a robust reward model. On the other hand, the \textit{Obj}=\textit{Subj} and \textit{Subj} systems require user feedback for every training dialogue as shown by the dashed orange line. 

Of course, the \textit{off-line RNN} system required no user feedback at all when training the system on-line since it had the benefit of prior access to a user simulator.  Its performance during training after the first 300 dialogues was, however, inferior to the {\it on-line GP} system.

\subsection{Dialogue Policy Evaluation} \label{sec:dpe}

In order to compare performance, the averaged results obtained between 400-500 training dialogues are shown in the first section of Table \ref{table:eval} along with one standard error. 
For the 400-500 interval, the 
\textit{Subj}, \textit{off-line RNN} and {\it on-line GP} systems achieved comparable results without statistical differences. The results of continuing training on the \textit{Subj} and {\it on-line GP} systems from 500 to 850 training dialogues are also shown.  As can be seen, the {\it on-line GP} system was significantly better presumably because it is more robust to erroneous user feedback compared to the \textit{Subj} system.

\begin{table}[h]
\caption{{Subjective evaluation of the \textit{Obj=Subj}, \textit{off-line RNN}, \textit{Subj} and \textit{on-line GP} system during different stages of on-line policy learning. \textit{Subjective}: user binary rating on dialogue success. Statistical significance was calculated using a two-tailed Student’s t-test with p-value of 0.05.}}
\begin{center}
\begin{tabular}{c|c|c}
{\bf Dialogues} & {\bf Reward Model}    & \textit{\textbf{Subjective} (\%)}   \\ \hline \hline
\multirow{4}{*}{400-500} & \textit{Obj=Subj}   & 85.0 $\pm$ 2.1    \\ %\cline{2-3}
& \textit{off-line RNN} &  89.0 $\pm$ 1.8 \\ %\cline{2-3}
& \textit{Subj} &  90.7 $\pm$ 1.7 \\ %\clfine{2-3}
& \textit{on-line GP} &  91.7 $\pm$ 1.6 \\ \hline %\hline
\multirow{2}{*}{500-850} & \textit{Subj}   & 87.1 $\pm$ 1.0    \\ %\cline{2-3}
& \textit{on-line GP} & {\bf 90.9 $\pm$ 0.9*} \\ \hline %\hline
\multicolumn{3}{l}{\it * $p < 0.05$} \\
\end{tabular}
\end{center}
\label{table:eval}
\end{table}

\subsection{Reward Model Evaluation} \label{sec:rml}

The above results verify the effectiveness of the proposed reward model for policy learning. Here we investigate further the accuracy of the model in predicting the subjective success rate. 
An evaluation of the \textit{on-line GP} reward model between 1 and 850 training dialogues is presented in Table \ref{tab:fmeasure}.

Since three reward models were learnt each with 850 dialogues, there were a total of 2550 training dialogues. Of these, the models queried the user for feedback a total of 454 times, leaving 2096 dialogues for which learning relied on the reward model's prediction.
The results shown in the table are thus the average over 2096 dialogues. 

As can be seen, there was a significant imbalance between success and fail labels since the policy was improving along with the training dialogues. This lowered the recall on failed dialogue prediction as the model was biased to data with positive labels. Nevertheless, its precision scores well.
On the other hand, the successful dialogues were accurately predicted by the proposed model.

\begin{table}[h]
\centering
\caption{Statistical evaluation of the prediction of the \textit{on-line GP} systems with respect to \textit{Subj} rating.}
\label{tab:fmeasure}
\begin{tabular}{c|cccc}
 \textit{\textbf{Subj}}                   & {\bf Prec.}                 & {\bf Recall} & {\bf F-measure} & {\bf Number} \\ \hline
Fail & 1.00 & 0.52 & 0.68      & 204         \\ 
Suc. & 0.95 & 1.00 & 0.97      & 1892        \\ \hline
Total               & 0.96                      & 0.95                      & 0.95      & 2096        \\ 
\end{tabular}
\end{table}

\subsection{Example Dialogues}  \label{app:sample}

The key benefits of the \textit{on-line GP} reward model compared to other models are its robustness to noise and efficient use of user supervision.
Since the four systems compared above differ only in the design of reward model (learning objective), their on-line behaviours were broadly similar.

\begin{table*}[t]
  \caption{Example dialogues between on-line users and the proposed {\it on-line GP} system.}
  \label{tab:sample}
  \centering
  \resizebox{2\columnwidth}{!}{
  \begin{tabular}{ll}
\toprule
%ACL2016 & GPReward2 $voip-2015671761-160216_010730$ CLEAN\\
\multicolumn{2}{l}{\bf -Low noise condition- (User Subjective Evaluation: {\it Success}, on-line GP Evaluation: {\it Success})}\\
Machine	&: Hello, welcome to the Cambridge dialogue system, What kind of food would you like?\\
Human	&: [Top ASR] Yeah I want to find a restaurant that serves European food.\\ 
		&\ \ [Top SEM] inform(food=european,type=restaurant)\\
Machine	&: Hotel du Vin and Bistro is a nice place. It serves European food.\\
Human	&: [Top ASR] Uh what is the address and phone number?\\
	&\ \ [Top SEM] request(phone,addr)\\
Machine	&: Hotel du Vin and Bistro is a nice place. Their address is 15-19 Trumpington Street and their phone is 01223 227330.\\
Human	&: [Top ASR] Thank you goodbye.\\
	&\ \ [Top SEM] bye()\\
Machine	&: Thank you for using this system.\\    
\midrule
%ACL2016 & GPReward1 $voip-2192565470-160215_204354$ CLEAN\\
\multicolumn{2}{l}{\bf -Higher noise condition- (User Subjective Evaluation: {\it Success}, on-line GP Evaluation: {\it Success})}\\
Machine	&: Hello, welcome to the Cambridge dialogue system, What kind of food would you like?\\
Human	&: [Top ASR] And I want to find a expensive restaurant it should be in the south part of town.\\ 
		&\ \ [Top SEM] inform(area=south,pricerange=expensive,type=restaurant)\\
Machine	&: Gourmet burger kitchen is a nice place it is in the expensive price range.\\
Human	&: [Top ASR] And do you have something else?\\
	&\ \ [Top SEM] null()\\
	&\ \ [2nd SEM] reqalts()\\
Machine	&: Did you say you are looking for a venue in the area of south ?\\
Human	&: [Top ASR] Yes.\\
	&\ \ [Top SEM] affirm()\\
Machine	&: Frankie and Benny's is a nice place, it is in the south part of town and it is in the expensive price range.\\
Human	&: [Top ASR] Okay can I have the uh address phone number called for any.\\
	&\ \ [Top SEM] request(addr)\\
	&\ \ [2nd SEM] request(phone,addr)\\
Machine	&: Frankie and Benny's is a nice place. Their address is Clifton Way and their phone number is 01223 412430.\\    
Human	&: [Top ASR] What is the pub called for free you any?\\
	&\ \ ...\\
	&\ \ [5th ASR] What is the post code for free you any?\\
	&\ \ [Top SEM] request(postcode)\\
Machine	&: Frankie and Benny's is a nice place. Their postcode is C.B 1, 7 D.Y.\\   
Human	&: [Top ASR] Thank you goodbye.\\
	&\ \ [Top SEM] bye()\\
Machine	&: Thank you for using this system.\\    
\bottomrule
  \end{tabular}}
\end{table*}

Two example dialogues between users and the \textit{on-line GP} system are listed in Table \ref{tab:sample} to illustrate how the system operates under different noise conditions. The user's subjective rating and the rating determined by the \textit{on-line GP} reward model are also shown.
The labels `{\it n}-th ASR' and `{\it n}-th SEM' indicate the n-th most likely hypotheses from speech recogniser and semantic decoder respectively.

\section{Conclusion} \label{sec:conclude}
In this paper we have proposed an active reward learning model using Gaussian process classification and an unsupervised neural network-based dialogue embedding to enable truly on-line policy learning in spoken dialogue systems.
The system enables stable policy optimisation by robustly modelling the inherent noise in real user feedback and uses active learning to minimise the number of feedback requests to the user. We found that the proposed model achieved efficient policy learning and better performance compared to other state-of-the-art methods in the Cambridge restaurant domain. A key advantage of this Bayesian model is that its uncertainty estimate allows active learning and noise handling in a natural way.
The unsupervised dialogue embedding function required no labelled data to train whilst providing a compact and useful input to the reward predictor. Overall, the techniques developed in this paper enable for the first time a viable approach to on-line learning in deployed real-world dialogue systems which does not need 
a large corpus of manually annotated data or the construction of a user simulator.

Consistent with all of our previous work, 
the reward function studied here is focused primarily on task success.
This may be too simplistic for many commercial applications and
further work will be needed in conjunction with human interaction experts to identify and incorporate the extra dimensions of dialogue quality that will be needed to achieve the highest levels of user satisfaction.

\subsubsection*{Acknowledgments}

Pei-Hao Su is supported by Cambridge Trust and the Ministry of Education, Taiwan. This research was partly funded by the EPSRC grant EP/M018946/1 \textit{Open Domain Statistical Spoken Dialogue Systems}. The data used in the experiments is available
at \href{http://www.repository.cam.ac.uk/handle/1810/256020}{\textit{www.repository.cam.ac.uk/handle/1810/256020}}.

\bibliography{acl2016}
\bibliographystyle{acl2016}

\end{document}